\documentclass[letterpaper, 10 pt, conference]{ieeeconf}  % Comment this line out if you need a4paper

\IEEEoverridecommandlockouts                              % This command is only needed if 
\usepackage[urlcolor=red]{hyperref}
\usepackage{amsmath} % assumes amsmath package installed
\usepackage{amssymb}  % assumes amsmath package installed
\usepackage{graphicx}
\usepackage{float}
\usepackage{xcolor}
\usepackage{subfigure}
\usepackage{subcaption}  % 用于处理子图
\usepackage{graphicx}    % 用于插入图片
\usepackage{tikz}
\usepackage{booktabs}
\usepackage{multirow}
\usepackage{siunitx}
\usepackage{moreverb,url}
\title{\LARGE \bf
Optimizing Efficiency of Mixed Traffic through Reinforcement Learning: A Topology-Independent Approach and Benchmark
}
\usepackage{xcolor}

\definecolor{deepgreen}{RGB}{0,100,80}

\author{Chuyang Xiao$^{1*}$, Dawei Wang$^{2*}$, Xinzheng Tang$^{3}$, Jia Pan$^{2}$ and Yuexin Ma$^{1}$% <-this % stops a space
\thanks{*Joint first authors}% <-this % stops a space
\thanks{$^{1}$Chuyang Xiao and Yuexin Ma are with the School of Information Science and Technology, ShanghaiTech University, Shanghai, 201210, China}%
\thanks{$^{2}$Dawei Wang and Jia Pan are with Department of Computer Science and TransGP, The University of Hong Kong, Pokfulam, HK.}%
\thanks{$^{3}$Xinzheng Tang is with Department of Civil Engineering, The University of Hong Kong, Pokfulam, HK.}%
}

\begin{document}

\maketitle
\thispagestyle{empty}
\pagestyle{empty}

%%%%%%%%%%%%%%%%%%%%%%%%%%%%%%%%%%%%%%%%%%%%%%%%%%%%%%%%%%%%%%%%%%%%%%%%%%%%%%%%
\begin{abstract}

This paper presents a mixed traffic control policy
designed to optimize traffic efficiency across diverse road
topologies, addressing issues of congestion prevalent in urban
environments. A model-free reinforcement learning (RL) ap-
proach is developed to manage large-scale traffic flow, using
data collected by autonomous vehicles to influence human-
driven vehicles. A real-world mixed traffic control benchmark
is also released, which includes 444 scenarios from 20 countries,
representing a wide geographic distribution and covering a
variety of scenarios and road topologies. This benchmark
serves as a foundation for future research, providing a realistic
simulation environment for the development of effective policies.
Comprehensive experiments demonstrate the effectiveness and
adaptability of the proposed method, achieving better perfor-
mance than existing traffic control methods in both intersection
and roundabout scenarios. To the best of our knowledge, this
is the first project to introduce a real-world complex scenarios
mixed traffic control benchmark. Videos and code of our work
are available at \url{https://sites.google.com/berkeley.edu/mixedtrafficplus/home}

\end{abstract}

%%%%%%%%%%%%%%%%%%%%%%%%%%%%%%%%%%%%%%%%%%%%%%%%%%%%%%%%%%%%%%%%%%%%%%%%%%%%%%%%
\section{INTRODUCTION}

Traffic serves as the lifeline of cities, fueling economic growth and ensuring the seamless execution of daily activities. Nevertheless, some issues such as traffic congestion can impede urban functionality, resulting in financial setbacks and inconveniences for commuters. Though there are numerous measures to mitigate traffic congestion, including infrastructure development and traffic signal optimization, the annual losses attributed to these issues continue to be substantial. 
As urban development accelerates, the escalating number of private vehicles and city construction further exacerbate the congestion problems. 
Moreover, relying solely on traffic signals for control proves to be unreliable, as extreme weather conditions or energy shortages may result in signal failure. In the absence of alternative coordination systems, such failures could lead to traffic gridlock and severe congestion. 

In the past decades, research works on traffic control by leveraging connected and autonomous vehicles (CAVs), have offered new opportunities~\cite{spielberg2019neural,pek2020using, feng2023dense}. Although these works demonstrated the possibility of controlling road traffic without conventional traffic light infrastructure, the universal connectivity and centralized control of all autonomous vehicles is still a too strong assumption that can not be realized in the near future. Fortunately, recent works on boosting road safety and efficiency under mixed traffic conditions provide new ideas and directions. For instance, Yan and Wu~\cite{yan2021reinforcement} proposed a system that integrates human-driven vehicles (HVs) and robot vehicles (RVs), using the behavior of RVs to influence nearby HVs, thereby replacing traffic signals to optimize intersection efficiency. Wang et al.~\cite{wang2023learning} introduced a model-free reinforcement learning (RL) framework that trains a policy to achieve the same control effects as traffic signals at real-world complex intersections. However, these approaches are explicitly designed for simple road topologies, restricting the adaptability of these frameworks to other complex real-world scenarios, including multi-legged intersections and roundabouts. 

\begin{figure}[H]
    \centering
    \begin{minipage}{1\linewidth} % 总宽度的一半
        \centering
        % 第一行的三张图
        \begin{minipage}{0.32\linewidth} % 每张图占总宽度的三分之一
            \centering
            \includegraphics[width=\linewidth, height=0.7\linewidth]{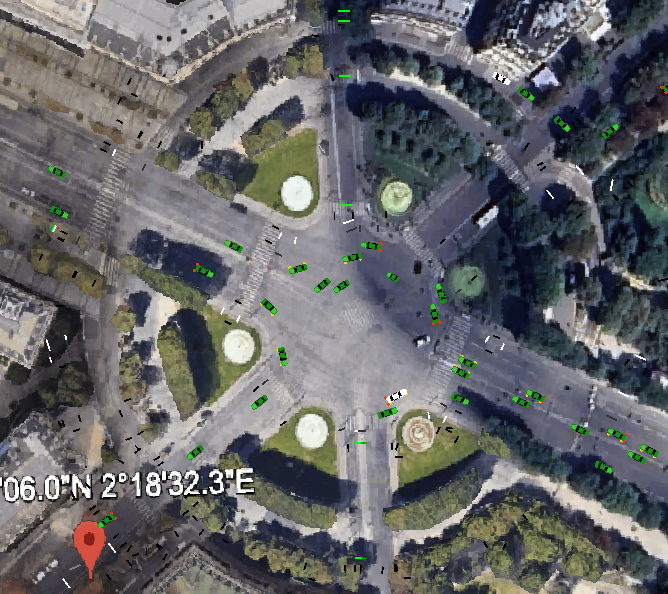}
        \end{minipage}%
        % \hspace{0.02\linewidth} % 控制图片之间的间距
        \begin{minipage}{0.32\linewidth}
            \centering
            \includegraphics[width=\linewidth, height=0.7\linewidth]{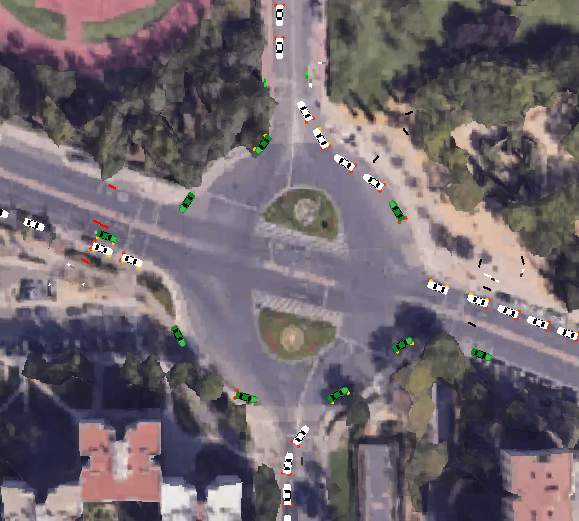}
        \end{minipage}%
        % \hspace{0.02\linewidth}
        \begin{minipage}{0.32\linewidth}
            \centering
            \includegraphics[width=\linewidth, height=0.7\linewidth]{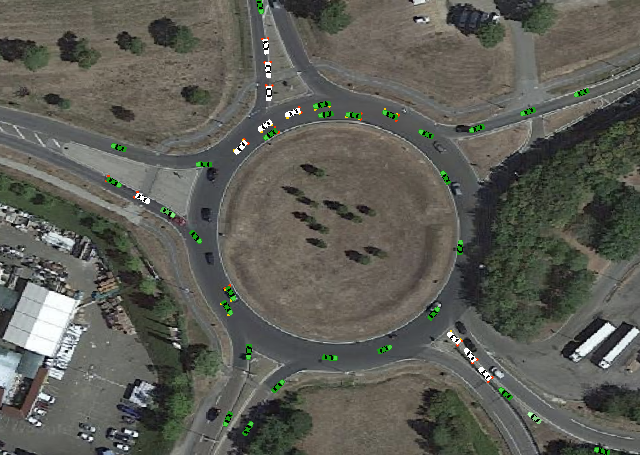}
        \end{minipage}

        % % 空行用于分隔两行
        % \vspace{0.2cm}

        % 第二行的三张图
        \begin{minipage}{0.32\linewidth} % 每张图占总宽度的三分之一
            \centering
            \includegraphics[width=\linewidth, height=0.7\linewidth]{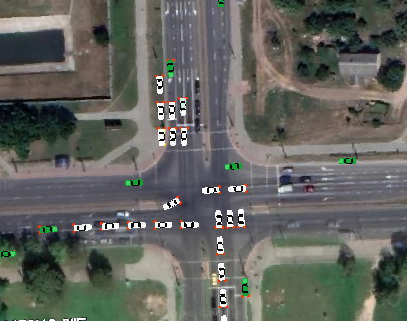}
        \end{minipage}%
        % \hspace{0.02\linewidth}
        \begin{minipage}{0.32\linewidth}
            \centering
            \includegraphics[width=\linewidth, height=0.7\linewidth]{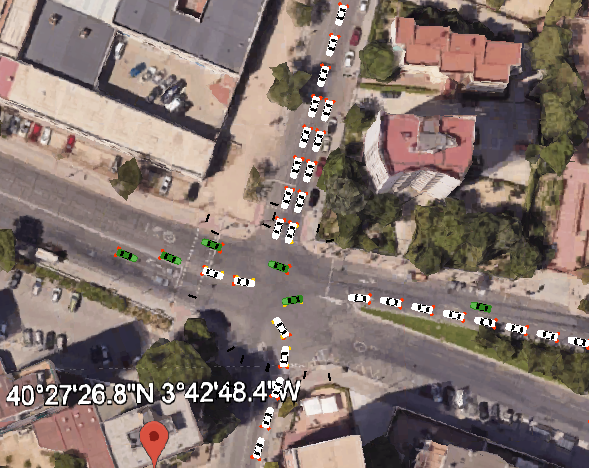}
        \end{minipage}%
        % \hspace{0.02\linewidth}
        \begin{minipage}{0.32\linewidth}
            \centering
            \includegraphics[width=\linewidth, height=0.7\linewidth]{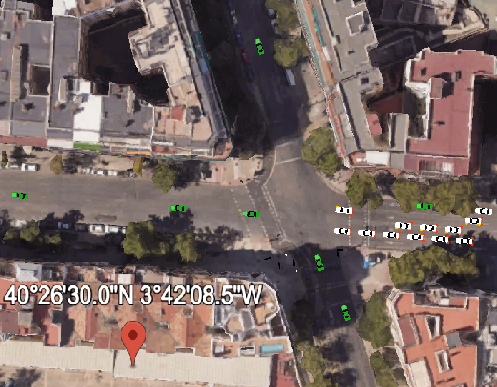}
        \end{minipage}
    \end{minipage}
    \caption{Benchmark Scenarios Visualization}
    \label{fig:figure1}
        \vspace{-1.0em}
\end{figure}

In this paper, our objective is to develop a mixed traffic control policy to optimize traffic efficiency, without constraint of particular topologies. To achieve this goal, we present a real-world mixed traffic control benchmark, covering hundreds of intersectional scenarios with wide-range road topologies, including intersections and roundabouts with various shapes, all over the world. Some of the road topologies investigated in this paper are illustrated in Fig.~\ref{fig:figure1}. In total, our benchmark includes 444 scenarios, collected from 20 countries around the world. As shown in Table~\ref{tab:benchmark_comparison}, we compared the scenario dataset used in our work and other mixed traffic control methods. It shows that our dataset contains the most complex and diverse real-world scenarios. To the best of our knowledge, our project is the first real-world complex scenarios mixed traffic control benchmark.
\begin{table*}[h!]
\centering
\resizebox{\textwidth}{!}{ % 自动调整表格宽度以适应页面
\begin{tabular}{c|c|c|c|c|c|c|c|c}
\hline
Methods                     & Scenarios \# & Real-world & Scenario Categories & Incoming lanes \# & Outgoing lanes \# & Legs \# & Traffic demand (veh/hr) & Countries \# \\
\hline
Jang~\cite{jang2019simulation}  & 1   & No    & Roundabout   & 2         & 2         & 4       & 720          & 0   \\
Yan~\cite{yan2021reinforcement} & 3   & No    & Intersection   & [3,6]     & [3,6]     & [4,12]  & [400,1000]   & 0   \\
Wang~\cite{wang2023learning}  & 9   & Yes   & Intersection   & [10,21]  & [3,14]  & [3,4]   & [900,1200]   & 1   \\
Flow~\cite{wu2021flow}   & 13  & No    & Both   & [1,2]     & [1,2]     & [2,4]   & [500,3000]   & 0   \\
% \hline
\textbf{Ours}                 & \textbf{444} & \textbf{Yes} & \textbf{Both} & \textbf{[1,25]} & \textbf{[2,23]} & \textbf{[3,13]} & \textbf{[400,5000]} & \textbf{20} \\
\hline
\end{tabular}
}

\caption{\label{tab:benchmark_comparison}A detailed comparison of mixed traffic control datasets and benchmarks.}
        \vspace{-1.5em}
\end{table*}

To deal with real-world complex traffic scenarios, we introduced a model-free RL approach to manage large-scale traffic flow at various road topologies. In our method, each RV collects data about its surrounding vehicles within and around the intersection via its local perception system and then encodes the data into fixed-length representations. Finally, the policy leverages the input representations and determines the RV's acceleration, deciding whether to accelerate or decelerate and by how much. Our approach falls into the centralized training and decentralized execution paradigm (CTDE). During the training process, each RV optimizes the global reward and updates the shared policy. While during execution, all RVs make independent decisions and collectively ensure smooth traffic flow, without the need for explicit coordination between traffic participants.

To evaluate our proposed policy, we compared our method with traffic light control method (TL) and traffic light absence baseline (NoTL), since there is no previous work that can work through all test sets in our benchmark. Our results showed that our trained policy outperforms the no-traffic-signal baseline, performs comparably to traffic signal control and consistently observes high throughput and low wait time, across all test scenarios with various topologies.

To further compare our proposed method with previous methods, we divided our test scenarios into two subsets, the intersection subset and the roundabout subset. We compared our policy with the state-of-the-art intersectional traffic control policy~\cite{wang2023learning} on the intersection subset. The experiment results demonstrated that our policy achieves better performance, which could successfully avoid intersection congestion and achieve better traffic efficiency than Wang~\cite{wang2023learning}. In addition, we also provided a comparison with the state-of-the-art traffic control method at roundabout topologies~\cite{jang2019simulation}. The results showed that our proposed method could consistently reduce the average waiting time of all traffic participants, and also surpass all baseline methods. 

Our work demonstrates the feasibility of controlling unsignalized mixed traffic across all real-world road topologies. The main contributions of this paper can be summarized as
% \textcolor{magenta}{Have added some contributions below.}
\begin{itemize}
    \item We developed a model-free approach to control and coordinate large-scale traffic at intersections with diverse topologies, demonstrating the model’s generalizability in maintaining high throughput.

    \item A benchmark encompassing various topologies for mixed traffic control has been established. This benchmark will serve as a foundation for future research, facilitating the generation of large-scale, realistic data to provide RVs with an enriched simulation environment for learning effective policies. To the best of our knowledge, our project is the first real-world complex scenarios mixed traffic control benchmark.

    \item We conducted several comprehensive experiments to evaluate our proposed method, including comparisons with state-of-the-art methods. The results illustrated the effectiveness, generalizability and adaptability of our proposed method.
\end{itemize}

\section{Related Works}

At present, traffic efficiency optimization techniques can be classified into two main categories: traditional traffic signal control and traffic control without the use of traffic lights.

Conventional traffic signal control methods~\cite{jacome2018survey, wei2019survey, mohamed2022traffic} have been widely accepted and deployed in the real world for optimization of intersectional traffic efficiency. Nevertheless, traffic signals can prove to be unreliable under certain circumstances.

Recently, there has been a surge in research works that focus on unsignalized intersections. This new stream of research leverages the capabilities of autonomous vehicles and connected vehicles. For instance, autonomous intersection management (AIM) methods typically coordinate all vehicles approaching an intersection towards a common objective~\cite{sharon2017protocol,miculescu2019polling}. However, these methods often necessitate centralized control over all autonomous vehicles, an assumption that may be overly strong in the near future.

Mixed traffic, denoting a scenario where human drivers and autonomous vehicles share the road simultaneously, is a feasible solution for the foreseeable future. Traditional mathematical approaches, such as defining and solving an optimization or control problem, are common solutions presented for mixed traffic control problems~\cite{wang2019controllability, karimi2020cooperative, cai2020summit, dai2021towards, wang2023general, lu2023cooperative, hickert2023cooperation}. Nevertheless, traditional methods usually necessitate explicit modeling of the system's traffic flow or fail to fully encompass the underlying dynamics of the traffic. On the other hand, reinforcement learning (RL) has achieved notable success in various traffic control tasks~\cite{li2023survey} without an explicit dynamic model. For example, Yan and Wu~\cite{yan2021reinforcement} utilized an RL method for traffic control and coordination, testing several simulated scenarios. Yan et al.~\cite{yan2021courteous} proposed an RL method to optimize mixed traffic flow at three-way intersections. Moreover, Wang et al.~\cite{wang2023learning} presented a model-free RL framework that trains a decentralized policy to effectively control and coordinate complex 4-legged intersections. However, these approaches are generally tailored for specific intersection topologies and may not be applicable to diverse scenarios, such as roundabouts or other intricate intersection designs.

In addition to intersections, several studies also address a variety of road topologies, e.g., ring roads, figure-eight roads~\cite{wu2021flow}, highway bottleneck and merge~\cite{vinitsky2018lagrangian,feng2021intelligent, wei2019mixed}, two-way intersections~\cite{yan2021reinforcement, villarreal2023hybrid, Villarreal2024Eco}, road networks~\cite{wang2023large,chinchali2019multi} and roundabouts~\cite{jang2019simulation}. Nevertheless, these scenarios often lack real-world complexity and cannot be directly implemented in general open-world situations.

In summary, the development of an effective policy that can perform consistently across all road topologies remains an open challenge.

\begin{figure}[h!]  
\includegraphics[width=\columnwidth]{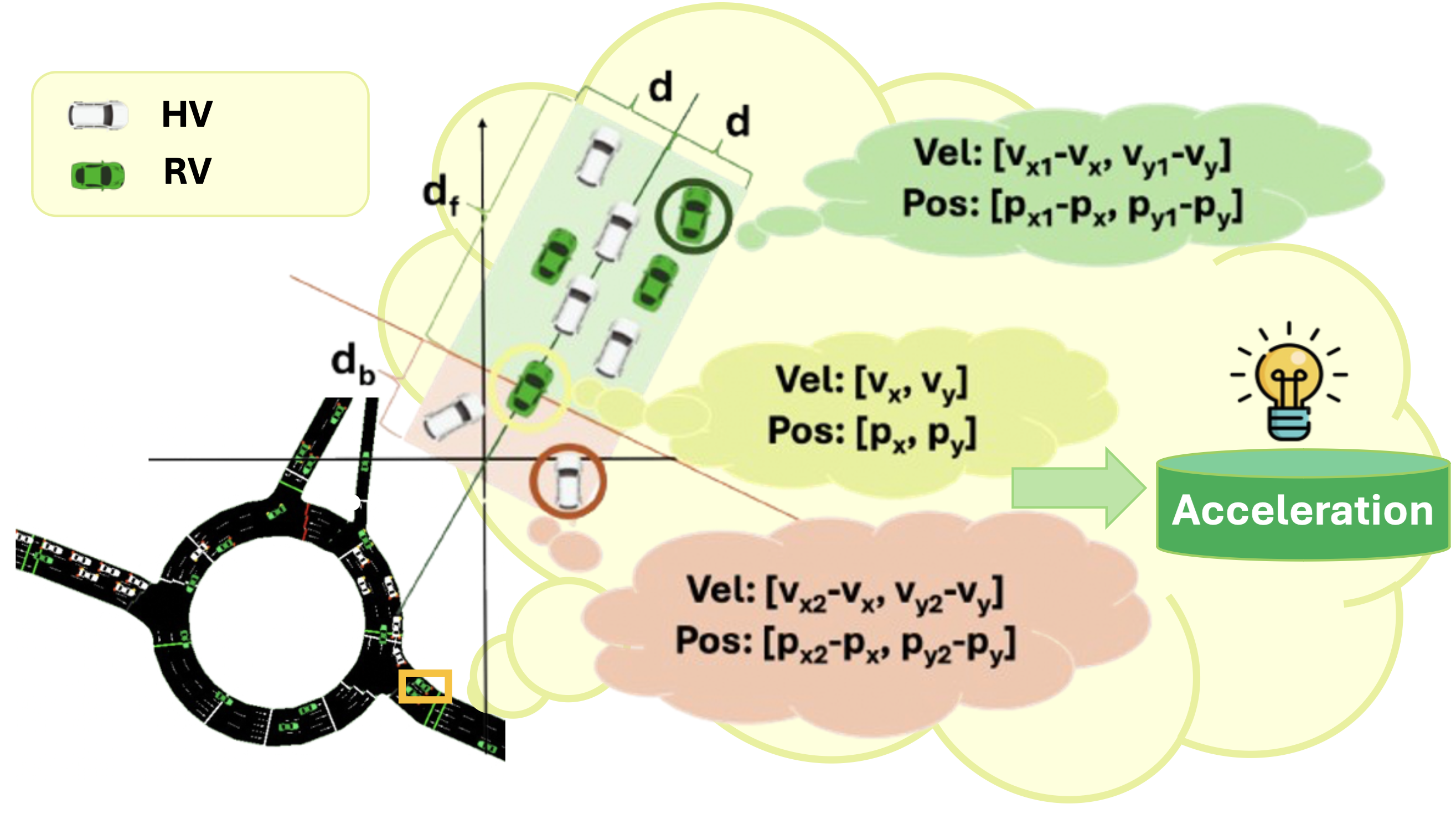}
\caption{
The pipeline of our method is illustrated as follows. LEFT: The green car within the yellow square represents the ego robot vehicle (RV), which independently collects traffic information via its local perception system and autonomously decides its acceleration. RIGHT: The policy observation is depicted. 
The vehicles surrounding the ego RV may be either robot vehicles (RVs: Green) or human-driven vehicles (HVs: White), but they will all be considered in the observation of the ego. Our method accounts for vehicles positioned both in front of the ego RV, represented by the light green area, and behind it, represented by the light red area. For each observed vehicle, its relative velocity and position to the ego vehicle are encoded into the observation. Here, for example, the central RV will include the HV behind it, located within the red circle, and the RV in front of it, located within the green circle, as part of its observation.
}
\label{fig:pipeline}
\end{figure}

\section{Methodology}
\subsection{Overview}
As depicted in Fig.\ref{fig:pipeline}, we present the detailed pipeline of our method. 
In contrast to previous works tailored to specific road topologies, our method is capable of adapting to a range of complex road configurations due to its topology-independent design. At each time step, every RV in our system determines its acceleration by considering the states of surrounding vehicles within a predefined distance range. The observation is conducted via the local perception system of the ego vehicle without communication and coordination with other traffic participants.

\subsection{Decentralized RL} 
We formulated mixed traffic control as a POMDP, which consists of a 7-tuple \((\mathcal{S}, \mathcal{A}, \mathcal{T}, \mathcal{R}, \Omega, \mathcal{O}, \gamma)\), where \(\mathcal{S}\) is a set of states (\(s \in \mathcal{S}\)), \(\mathcal{A}\) is a set of actions (\(a \in \mathcal{A}\)), \(\mathcal{T}\) is the transition probabilities between states \(T(s' \mid s, a)\), \(\mathcal{R}\) is the reward function \((\mathcal{S} \times \mathcal{A} \rightarrow \mathbb{R})\), \(\Omega\) is a set of observations \(o \in \Omega\), \(\mathcal{O}\) is the set of conditional observation probabilities, and \(\gamma \in [0, 1)\) is a discount factor. At each time \(t\), when an RV \(i\) enters the control zone, its action \(a^t_i\) is determined based on the current traffic condition \(o^t_i\), which is a partial observation of the traffic state \(s^t_i\) around the ego vehicle. We presented the policy \(\pi_\theta\) as a neural network trained using the following loss: \begin{equation} 
\left( R_{t+1} + \gamma_{t+1} q_{\bar{\theta}} \left( S_{t+1}, \arg \max_{a'} q_\theta (S_{t+1}, a') \right) - q_\theta (S_t, A_t) \right)^2, 
\end{equation} 
where \(q\) denotes the estimated value from the value network, while \(\theta\) and \(\bar{\theta}\) respectively represent the value network and the target network~\cite{hessel2018rainbow}. The target network is a periodic copy of the value network, which is not directly optimized during training.

\subsubsection{Action Space}
Due to the complexity of mixed traffic scenarios, optimizing both longitudinal and lateral acceleration simultaneously remains a challenging task. This is evidenced by previous works~\cite{wang2023learning, jang2019simulation}, which focus exclusively on longitudinal control. Furthermore, lateral control involves discrete lane change decisions, making it difficult for reinforcement learning (RL) algorithms to optimize continuous and discrete actions concurrently~\cite{delalleau2019discrete}. Therefore, in this study, we focused solely on acceleration as the action, while lane-changing decisions are governed by the simulator's default mechanism~\cite{erdmann2015sumo}. The action space is continuous, ranging between $[\SI{-10}{m/s^2}, \SI{10}{m/s^2}]$, and this acceleration is applied within the vehicle's current lane. 
The vehicle speed is restricted to non-negative values to prohibit the vehicle from reversing, and it must not surpass the default maximum lane speed to adhere to speed limitations.

\subsubsection{Observation Space}
The primary motivation of our method is to ensure adaptability to various road topologies. 
Consequently, each vehicle's observation is confined to the states of vehicles within a fixed observation area surrounding it. This observation includes the relative velocity and position of nearby vehicles to the central vehicle, collected solely through the vehicle's local perception system.

As shown in Fig.~\ref{fig:pipeline}~RIGHT, the observed area can be divided into two parts, the front area and the rear area. The front observation area, which is shown in light green, where $d_f$, refers to the maximum observation distance ahead of the ego RV, and $d$ indicates the maximum observation distance on both sides of the ego RV. The rear observation area, which is shown in light red, where $d_b$ represents the maximum observable distance behind the ego RV. 
This information is processed to create a fixed-length representation of the local traffic context. 

In this work, we chose $d_f = \SI{50}{m}$, $d_b = \SI{20}{m}$, and $d = \SI{5}{m}$ based on empirical knowledge. These distances are reasonable and align with the visual range typically observed while driving, as they encompass the visible distances on two adjacent lanes as well as the front and rear visibility ranges. The width is kept narrow enough to exclude irrelevant lanes, and the length is sufficient to ensure that vehicles can observe both the front and back effectively. We further limited the maximum number of observable vehicles to $N_f = 10$ nearest vehicles within the front observation area and $N_b = 5$ nearest vehicles within the rear observation area. These numbers are derived from the average vehicle count observed in these areas through repeated simulation experiments.  
Finally, our observation space is a continuous vector space with a size of  $(N_f + N_b) * 4$. Our policy then utilizes this observation to compute the optimal acceleration $a^t_i$ for each RV $i$, ensuring that the vehicle navigates safely and efficiently within the dynamic traffic environment. 

\begin{figure}[h!]  
\includegraphics[width=\columnwidth]{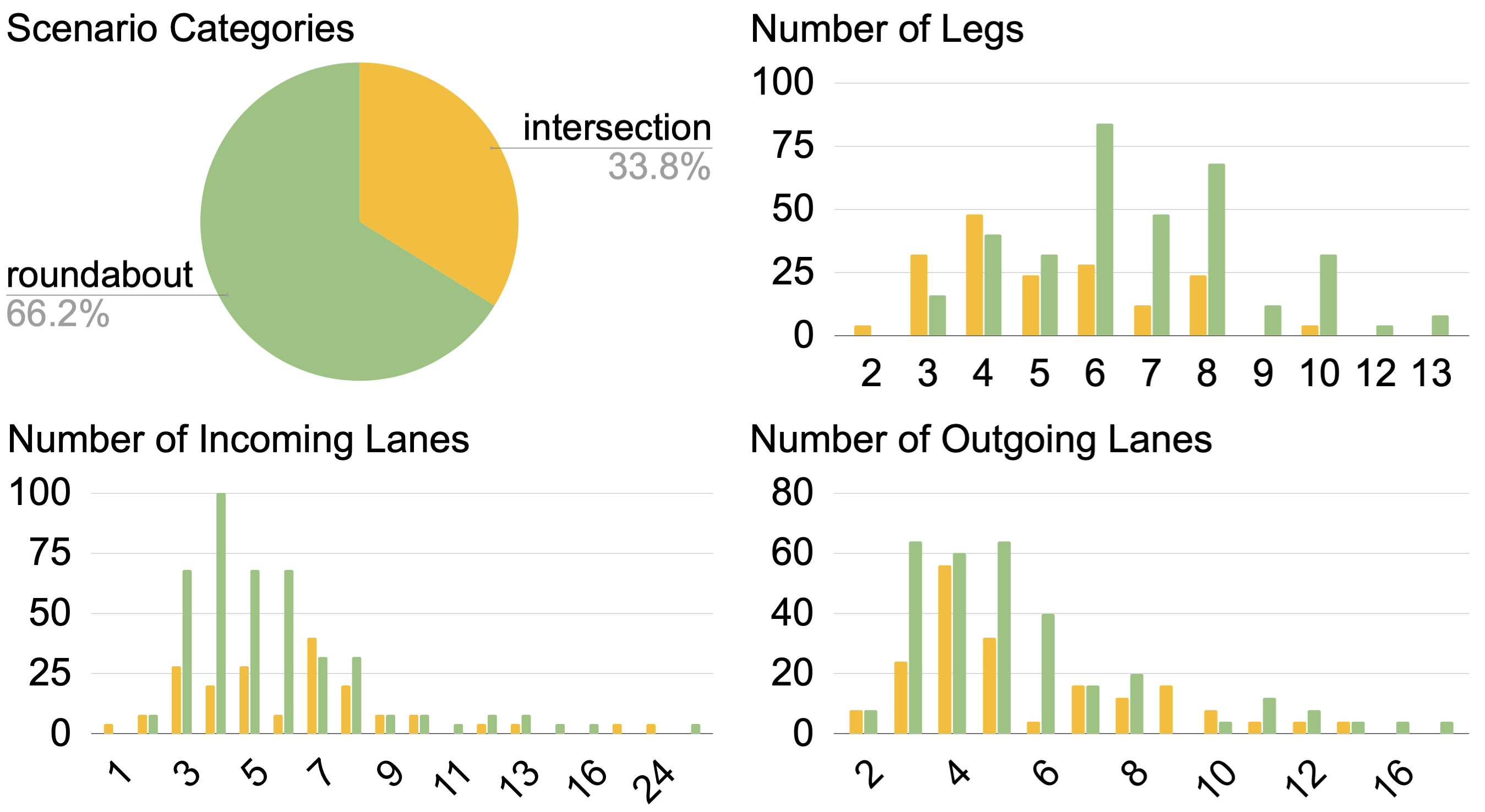}
\caption{Our scenario dataset is categorized by major topology types, intersections, and roundabouts. The scenarios are classified by the road topologies based on the number of road legs, the number of incoming lanes, and the number of outgoing lanes, to illustrate the distribution of the dataset. }

\label{fig:benchmark_overview}
\end{figure}
\subsubsection{Reward Function}
To encourage the RV to not only boost the overall traffic efficiency but also ensure safety, we designed a reward function $R$ as following:
\begin{equation}
\text{$R$} = \text{$\alpha*R_{Throughput} + \beta*R_{Collision} + \gamma*R_{Wait}$}
\end{equation}

The reward function $R$ consists of three components. The first term \(R_{\text{Throughput}}\), which is designed to increase the total traffic throughput. The total throughput equals the number of vehicles that have passed through the intersection/roundabout within the current simulation step.

The second term \(R_{\text{Collision}}\), which aims to minimize the collision behaviors and pursue safety, is formulated as: 
\begin{equation}
\text{$R_{Collision}$} = 
\begin{cases} 
1 & \text{if } \text{collision number} < 1 \\
-\text{collision number} & \text{otherwise} 
\end{cases}
\end{equation}
where the collision number refers to the number of vehicles that have experienced collisions within the current simulation step.

The last term \(R_{\text{Wait}}\), which encourages the policy to minimize the overall average waiting time of all vehicles in the scenario. It is defined as follows: 
\begin{equation}
R_{\text{Wait}} = 
\begin{cases} 
1 & \text{if } W_l \leq \text{W} \leq W_h \\
-\frac{\left|\text{W} - \frac{W_l + W_h}{2}\right|}{\frac{W_l + W_h}{2}} & \text{otherwise} 
\end{cases}
\end{equation}
where $W$ represents the average waiting time of all vehicles in the simulation at a given time step. 
The waiting time for each vehicle can be directly obtained from the SUMO interface during the simulation. It is defined as the time (in seconds) that the vehicle has spent moving at a speed below \( 0.1 \, \text{m/s} \), and is reset to zero whenever the vehicle accelerates and its speed exceeds \( \SI{0.1}{m/s} \) again. $W_l$ and $W_h$ are the lower and upper bounds of waiting times, respectively. These bounds are determined by collecting the average waiting times of all vehicles when the scenario is controlled by traffic lights. 
The midpoint between these bounds is chosen as the ideal waiting time.
 
To guarantee stable training, each of these components is normalized to a range of \([-1, 1]\) before being multiplied by the respective coefficients. Specifically, to normalize the observation, each observed vehicle is classified as being in either the front or rear area relative to the central vehicle. For vehicles in the front region, the x and y coordinates of their relative location are normalized by dividing by half of the region's width or the region's length respectively. A similar approach is used for vehicles in the rear region. All vehicles' relative velocities are normalized by dividing by the maximum velocity. If the vehicle's speed exceeds the maximum speed, we capped the speed at this maximum value. 
If the quantity of vehicles is fewer than the predefined observable vehicle number, the observation is padded as follows: In the front area, a relative position and velocity of \([1, 1]\) are appended, indicating they are situated at the greatest distance from the ego vehicle and are moving away at the maximum speed. Similarly, in the rear area, a \([-1, -1]\) of relative position and velocity are incorporated into the observation. This information exerts minimal influence on the ego vehicle and ensures that the policy prioritizes proximate vehicles surrounding the ego RV, particularly those whose velocities significantly impact the ego vehicle. Conversely, if the number of vehicles exceeds the required amount, we sorted them by their distance from the ego vehicle and include only the closest ones in the observation vector. 

\subsection{RL Algorithm}
We employed the Soft Actor-Critic (SAC) algorithm~\cite{haarnoja2018soft}, which is a widely used off-policy reinforcement learning technique known for its stability and sample efficiency. It optimizes both the policy and value functions by minimizing the soft Bellman residual while maximizing the entropy of the policy, leading to more exploratory and robust policies. The RL policy is architectured as a neural network, incorporating two fully connected (FC) layers with 256 hidden units each, and ReLU activation functions. We employed a centralized training and decentralized execution paradigm in our method. The policy is subject to offline training, leveraging centralized information, and subsequently operates in a decentralized fashion during online execution.

\section{New Benchmark and Scenario sets}
Our benchmark consists of 111 different road topologies collected from 20 countries all over the world. The topologies include different kinds of intersections and roundabouts. To generate high-fidelity simulation scenarios, we added traffic flow with different traffic demand $\in [400, 5000]$ and then 444 dynamic scenarios are generated. We divided our scenarios dataset into two parts, 372 scenarios for training and 72 scenarios for testing.
\subsubsection{Map conversion and scenario generation}
To generate real-world scenarios, we utilized the traffic simulator Simulation of Urban Mobility (SUMO)~\cite{behrisch2011sumo}, a platform extensively adopted for traffic simulation. Initially, we downloaded a real-world map from Open Street Map\footnote{https://www.openstreetmap.org}, an open-source geographic database. The downloaded map file, in OSM format, undergoes processing and preparation for SUMO using a variety of tools. 

Primarily, the OSM file is transformed into a SUMO network via the netconvert tool, thereby reconstructing the comprehensive traffic plan based on the network. During this phase, any traffic light systems (TLSs) lacking a specified type are assigned the STR program, an acronym for "Static Timed Regulation". This is a traffic light control strategy where the timings are fixed and unresponsive to real-time traffic conditions. Subsequently, we utilized the duarouter tool\footnote{https://github.com/eclipse-sumo/sumo/blob/main/tools/randomTrips.py} to generate a series of random trips within the network. To maintain a realistic traffic flow, vehicles are generated solely at the network's starting edges, avoiding the middle of edges. Finally, the scenario is primed for importation into the SUMO simulator.
To assess the impact of policies, we disabled the traffic signals and eliminated the simulation's inherent right-of-way policy for vehicles. This implies that the decisions of RVs should be dictated by the control policy, without any prior knowledge of traffic regulations.

\subsubsection{Mixed Traffic Construction}
During the simulation, to incorporate varying penetration rates of RVs, the newly spawned vehicles will be assigned as RV or HV randomly according to a predefined probability $P_{rv}$. RVs operate based on acceleration commands derived from a shared RL policy. HVs, on the other hand, follow the Intelligent Driver Model (IDM)~\cite{kesting2010enhanced}, which is widely accepted to simulate human driving behaviors. 

\subsubsection{diversity of dataset}
In Fig.~\ref{fig:benchmark_overview}, we show the diversity of our benchmark. Our benchmark encompasses two major topology categories: intersections (33.8\%) and roundabouts (66.2\%). The benchmark features a total of 111 distinct road topologies and 444 dynamic scenarios, created by applying various traffic demands $\in [400, 5000]$. The diversity in our road topologies is evident from the wide range of configurations, including the number of road legs $\in [3,13]$, incoming lanes $\in[1,5]$, and outgoing lanes $\in [2,23]$. The scenario dataset exhibits a broad distribution, effectively reflecting the diverse conditions and scenarios encountered in the real world.

\section{Experiments and Results}
\subsection{Experiment Setup}
\begin{table}[h!]
\centering
\begin{tabular}{ll}
\hline
Parameters                  & Value               \\ \hline
$\alpha$   & $1$                   \\
$\beta$   & $2$                   \\
$\gamma$   & $5$                   \\
$a_{max}$               & $\SI{10}{m/s^2}$             \\
$a_{min}$               & $\SI{-10}{m/s^2}$           \\
prioritized replay buffer $\alpha$ & $0.5$                 \\
replay buffer capacity     & $5e4$               \\
hidden layers              & $[256, 256]$\\
discount factor            & $0.99$              \\
minibatch size              & $256$             \\ 
learning rate               & $3e-4$      \\ 
target smoothing coefficient & $5e-3$\\
$d_f$                       & $\SI{50}{m}$ \\
$d_b$                       & $\SI{20}{m}$ \\
$d$                         & $\SI{5}{m}$ \\
$N_f$                       & $10$ \\
$N_b$                       & $5$ \\
$W_l$                       & $20$ \\
$W_h$                       & $30$ \\ 
% $P_{c2a}$                       & $0.5$ \\
% $P_c$                       & $0.5$ \\
% $P_a$                       & $0.5$ \\
$P_{rv}$                       & $\{0.4, 0.5. 0.7. 0.8, 0.9, 1.0\}$ \\\hline

\end{tabular}
\caption{Hyperparameters of our method}
        \vspace{-1.5em}
\label{tab:hyperparameters}
\end{table}
\begin{table*}[h!]
\centering
\begin{tabular}{c|cc|cc|cc}
\hline
\textbf{Test Set}       & \multicolumn{2}{c|}{\textbf{Whole Test Set}}                                                                                                                     & \multicolumn{2}{c|}{\textbf{Intersection Subset}}                                                                                                               & \multicolumn{2}{c}{\textbf{Roundabout Subset}}                                                                                                                   \\
\textbf{Metric}         & \textbf{\begin{tabular}[c]{@{}c@{}}Throughput \\ ($10^{-3}$)\end{tabular}} & \textbf{\begin{tabular}[c]{@{}c@{}}Avg. Wait \\ Time (s)\end{tabular}} & \textbf{\begin{tabular}[c]{@{}c@{}}Throughput\\ ($10^{-3}$)\end{tabular}} & \textbf{\begin{tabular}[c]{@{}c@{}}Avg. Wait \\ Time (s)\end{tabular}} & \textbf{\begin{tabular}[c]{@{}c@{}}Throughput \\ ($10^{-3}$)\end{tabular}} & \textbf{\begin{tabular}[c]{@{}c@{}}Avg. Wait \\ Time (s)\end{tabular}} \\ \hline
\textbf{NoTL}           & 1.74                                                                                   & 513.27                                                                 & 1.95                                                                                   & 588.61                                                                 & 1.53                                                                                    & 537.08                                                                 \\
\textbf{TL}             & 10.65                                                                                   & 503.06                                                                 & 13.40                                                                                  & 467.83                                                                 & 7.89                                                                                    & 205.21                                                                 \\
\textbf{Wang~\cite{wang2023learning}}   & --                                                                                      & --                                                                     & 16.99                                                                                  & 488.74                                                                 & --                                                                                      & --                                                                     \\
\textbf{Jang~\cite{jang2019simulation}}   & --                                                                                      & --                                                                     & --                                                                                     & --                                                                     & 3.29                                                                                    & 28.24                                                                  \\
\textbf{Ours (40\%RV)}  & 3.14                                                                                    & 388.24                                                                 & 3.65                                                                                   & 475.38                                                                 & 2.62                                                                                    & 302.08                                                                 \\
\textbf{Ours (50\%RV)}  & 4.74                                                                                    & 321.93                                                                 & 5.64                                                                                   & 435.86                                                                 & 3.84                                                                                    & 178.95                                                                 \\
\textbf{Ours (60\%RV)}  & 5.89                                                                                    & 290.49                                                                 & 7.16                                                                                   & 433.38                                                                 & 4.61                                                                                    & 161.21                                                                 \\
\textbf{Ours (70\%RV)}  & 6.56                                                                                   & 280.28                                                                 & 7.35                                                                                   & 403.95                                                                 & 5.77                                                                                    & 119.00                                                                 \\
\textbf{Ours (80\%RV)}  & 9.25                                                                                   & 222.62                                                                 & 11.60                                                                                  & 332.13                                                                 & 6.90                                                                                    & 110.09                                                                 \\
\textbf{Ours (90\%RV)}  & 12.11                                                                                   & 204.06                                                                 & 15.33                                                                                  & 319.52                                                                 & 8.89                                                                                    & 74.11                                                                  \\
\textbf{Ours (100\%RV)} & \textbf{17.5}                                                                           & \textbf{130.79}                                                        & \textbf{20.50}                                                                         & \textbf{183.17}                                                        & \textbf{14.5}                                                                           & \textbf{9.11}                                                         \\ \hline
\end{tabular}
\caption{The overall results measured in average wait time (s) and throughput rate ($10^-3$) between our method and four baseline methods on the whole test set, intersection subset and roundabout subset. In instances where a method was inapplicable to certain test sets, the corresponding result was left blank.}
        \vspace{-1.5em}
\label{tab:evaluation_test_set}
\end{table*}
% \textcolor{red}{summarize parameters in a table if possible.}
In conducting the experiments, we trained our SAC algorithm utilizing NVIDIA A6000 GPUs (8 GPUs) and 64 CPUs, for 43 hours. The comprehensive set of hyperparameters employed in this study is elaborated in Table.~\ref{tab:hyperparameters}.

We implemented and trained the baseline methods on our training dataset. Since the Jang~\cite{jang2019simulation} is only applicable to roundabout scenarios, it was trained on the roundabout subset of the training set. Similarly, Wang~\cite{wang2023learning} was trained on the intersection subset. Our method was trained on the whole training set including roundabout and intersection.

\subsection{Overall Performance}

We tested our method with varying RV penetration rates, ranging from 40\% to 100\%. For both the roundabout and intersection scenarios, we used TL (traffic light control) and NoTL (traffic light absence) as baselines. Additionally, for the roundabout scenarios, we included the work of Jang et al~\cite{jang2019simulation} as the comparison method. On the intersection subset, we compared our method with state-of-the-art intersectional traffic control method~\cite{wang2023learning}.

Each evaluation is repeated five times and lasts 3000 simulation steps. Subsequently, the metrics derived from these tests were averaged to yield the final performance evaluation for each method. We depicted the throughput rate and average waiting time of our method, considering RV penetration rate ranging from 40\% to 100\%, and the other four comparison methods mentioned above. In instances where a method was inapplicable to certain test sets, the corresponding result was left blank.

The experimental results from our comprehensive benchmark, along with the intersection and roundabout subsets, are presented in Table~\ref{tab:evaluation_test_set}. These experiments demonstrated that with an increase in the RV rate, the throughput rate consistently ascends, while the average waiting time correspondingly descends. The best results were observed when the RV rate reached 100\%, at which point both metrics significantly outperformed the other three baselines. With 100\% RVs, our method was able to reduce the average waiting time by a notable 74\% and enhance the total throughput rate by 64\% in comparison to the traditional traffic light baseline. Moreover, our method consistently outperforms the conventional traffic signal control baseline in terms of average waiting time when the RV rate exceeds 40\%.

\begin{figure}[H]  
\includegraphics[width=\columnwidth]{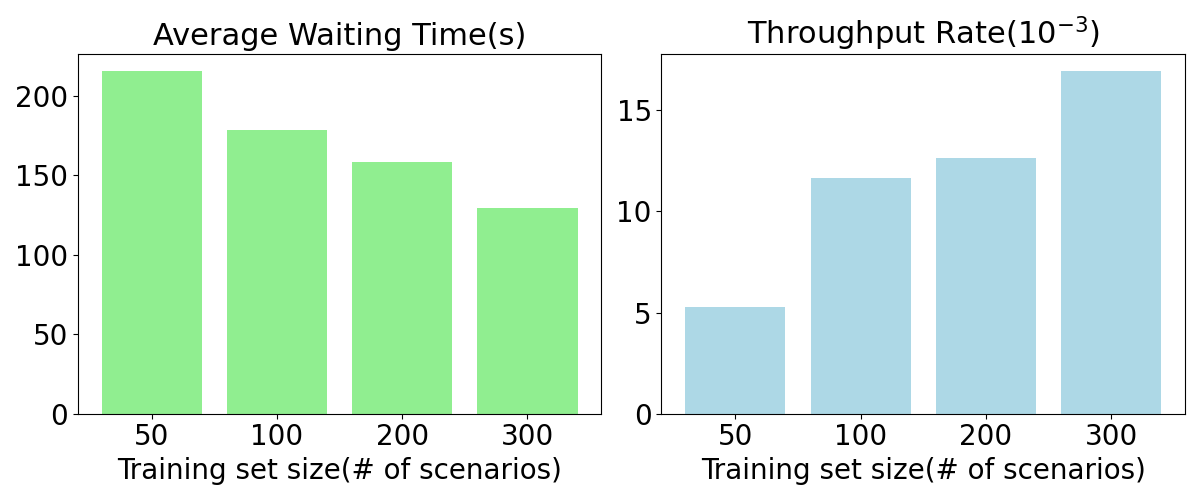}
    \caption{An evaluation of our policy's performance, with training set sizes varying from 50 to 300 scenarios, was conducted based on two key metrics: average waiting time (s) and throughput rate ($10^-3$). The results clearly demonstrate improvements in both metrics with an increase in the size of the training set. The decreasing average waiting time and increasing throughput rate indicate that our policy becomes more effective with the enlargement of the training set.}
% suggesting that further expansion of the training set could lead to even better performance.}
        % \vspace{-1.5em}

\label{fig:benchmark_size_test}
\end{figure}

To evaluate our method's performance change as the size of the training set scaled up, we constructed and trained four distinct models, each corresponding to 50, 100, 200, and 300 scenarios respectively. These models were then subjected to evaluation on a benchmark test set. As illustrated in Fig.~\ref{fig:benchmark_size_test}, there is a notable improvement in the policy's performance commensurate with an increase in the number of training scenarios. Specifically, an escalation in the training set from 50 to 300 scenarios led to nearly a 200\% increase in performance, as measured by the total throughput rate. Furthermore, the benchmark can be effortlessly expanded by acquiring further road topologies from the OpenStreetMap website or by integrating new traffic demands into topologies. This adaptability facilitates the continuous enhancement of our policy via the expansion of our benchmark's scope.

\subsection{Traffic Demands}
% \textcolor{red}{TODO.}
In Table~\ref{tab:case_study}, we evaluated our method with four baselines under two traffic demand levels, applicable to both intersection and roundabout topologies, utilizing the throughput rate metric ($10^{-3}$). In intersection scenarios, our method achieves the highest throughput at both 3000 and 5000 veh/hr, thereby demonstrating a significant superiority in managing elevated traffic demand in comparison to the baselines. As traffic demand escalates, the throughput of both the NoTL and TL methods witness a substantial decrease, signifying that congestion occurred at the intersection area. For roundabout scenarios, while all methods undergo a decrease in throughput rate in line with rising traffic demand, our approach consistently exceeds the performance of the baselines, particularly at higher demand levels, implying its robustness in mitigating traffic congestion. The results show that our method exhibits commendable performance even under extremely high traffic demands and continues to stably regulate the traffic stream.
% }

\begin{table}[]
\centering
\begin{tabular}{ccccc}
\hline
Scenario                & \multicolumn{2}{c}{Intersection} & \multicolumn{2}{c}{Roundabout} \\ \cline{2-5} 
Traffic Demand (veh/hr) & 3000            & 5000           & 3000           & 5000          \\ \hline
NoTL                    & 1.45            & 1.36           & 0.69           & 0.56          \\
TL                      & 11.71           & 6.61           & 7.11           & 3.80          \\
Jang~\cite{jang2019simulation}                    & --              & --             & 2.31           & 2.04          \\
Wang~\cite{wang2023learning}                    & 15.47           & 7.83           & --             & --            \\
Ours (100\%RV)          & \textbf{17.92}           & \textbf{18.67}          & \textbf{15.47}          & \textbf{10.43}         \\ \hline
\end{tabular}
\caption{
    We evaluated our proposed method along with four baseline methodologies, under two distinct levels of traffic demand and two types of road topologies. The throughput rate ($10^-3$) served as the evaluation metric in this assessment.}
% Throughput rate ($10^{-3}$) comparison between our method and three baselines under two traffic demand levels for both intersection and roundabout scenarios.}

        % \vspace{-1.5em}
\label{tab:case_study}
\end{table}

\section{CONCLUSIONS}

In this paper, we presented a novel model-free reinforcement learning (RL) approach designed to coordinate mixed traffic across a variety of complex and diverse road topologies. We showcased the efficiency and effectiveness of our method in sustaining high traffic throughput and minimizing wait time at unsignalized intersections and roundabouts. Our approach proved to be adaptable to varying situations and exhibited significant performance enhancements over existing traffic control methods under diverse traffic conditions. In addition, a comprehensive mixed traffic control benchmark comprising hundreds of scenarios, covering a wide spectrum of real-world road topologies, is released. To the best of our knowledge, our benchmark is the first real-world complex scenarios mixed traffic control benchmark. We anticipate that our benchmark, coupled with our innovative RL method, will serve as useful resources for the research community, encouraging further exploration and advancement in this area of study.

% \textcolor{magenta}{In this paper, we introduced a novel model-free reinforcement learning (RL) approach to coordinate mixed traffic across diverse and complex road topologies. We demonstrated the effectiveness of our method in maintaining high traffic throughput and low wait time, particularly in unsignalized intersections and roundabouts by introducing our robot vehicles to the scenarios. Our approach demonstrates to be adaptable to different scenarios, exhibit significant performance improvements over existing traffic control methods under varying traffic condition. Moreover, we released a comprehensive benchmark consisting of hundreds of scenarios from over 20 countries, covering a broad spectrum of real-world road topologies. We hope that our benchmark, along with the novel RL method, will serve as valuable resources for the research community to further build upon and advance this field of study.

% \addtolength{\textheight}{-12cm}

{\small
\bibliographystyle{IEEEtran}
\bibliography{references}
}

\end{document}